\documentclass[letterpaper, 10 pt, conference]{ieeeconf}  
\IEEEoverridecommandlockouts                              
\usepackage{graphicx} 
\usepackage{siunitx}
\usepackage{graphicx}
\usepackage{wrapfig}
\usepackage[flushleft]{threeparttable}
\usepackage[T1]{fontenc}
\usepackage{color}
\usepackage[bottom]{footmisc} 
\usepackage{enumerate}
\usepackage{multirow}
\usepackage{algpseudocode}
\usepackage{etoolbox}
\usepackage{amsmath}
\usepackage{algorithm}
\usepackage{algorithmicx}
\usepackage{graphicx}
\usepackage{caption}
\usepackage{subcaption}
\usepackage{soul}
\usepackage{xcolor,colortbl}
\usepackage[bookmarks=false]{hyperref}

\definecolor{Gray}{gray}{0.25}
\definecolor{red}{rgb}{1.00,0.00,0.00}
\definecolor{blue}{rgb}{0.00,0.00,1.00}
\definecolor{green}{rgb}{0.4,1.00,0.0}
\definecolor{yellow}{rgb}{0.5,0.5,0.0}

\newcommand{\cblue}[1] {\textcolor{blue}{#1}}

\title{{\small \vspace{-1cm} \cblue{Accepted in the 29th IEEE International Conference on Robot \& Human Interactive Communication (\textbf{RO-MAN 2020})}}\\ \vspace{0.5cm} Learning to Grasp 3D Objects using Deep Residual U-Nets}

\author{Yikun Li, Lambert Schomaker, S. Hamidreza Kasaei
\thanks{The authors are with the Faculty of Science and Engineering, Artificial Intelligence and Computer Science, University of Groningen, 9700 AB Groningen, The Netherlands.
{\tt\small\{yikun.li, l.r.b.schomaker, hamidreza.kasaei\}@rug.nl}}
\thanks{We are grateful to the NVIDIA corporation for supporting our research through the NVIDIA GPU Grant Program.}
}

\begin{document}

\maketitle
\thispagestyle{empty}
\pagestyle{empty}

\begin{abstract}
Grasp synthesis is one of the challenging tasks for any robot object manipulation task. In this paper, we present a new deep learning-based grasp synthesis approach for 3D objects. In particular, we propose an end-to-end 3D Convolutional Neural Network to predict the objects' graspable areas. We named our approach \emph{Res-U-Net} since the architecture of the network is designed based on U-Net structure and residual network-styled blocks. It devised to plan 6-DOF grasps for any desired object, be efficient to compute and use, and be robust against varying point cloud density and Gaussian noise. We have performed extensive experiments to assess the performance of the proposed approach concerning graspable part detection, grasp success rate, and robustness to varying point cloud density and Gaussian noise. Experiments validate the promising performance of the proposed architecture in all aspects. A video showing the performance of our approach in the simulation environment can be found at \href{http://youtu.be/5_yAJCc8owo}{\cblue{\footnotesize \texttt{http://youtu.be/5\_yAJCc8owo}}}
\end{abstract}

\section{Introduction}

Traditional object grasping approaches have been widely used in many industrial settings, such as factories assembly lines.  In such domains, robots broadly work in tightly controlled conditions to perform object manipulation tasks.  Nowadays, service robots are entering human-centric environments. In such unstructured places, generating stable grasp configuration for the household objects is challenging because of the high demand for precise and real-time response in unpredictable and fast-changing environmental conditions \cite{gibson2014ecological}. 
In human-centric environments, an object may have many graspable areas/points, where each one can be used to accomplish a specific task. 
As an example, consider a robotic cutting task using a knife. The knife has two graspable areas: the handle and the blade. The blade is used to cut through material, and the handle is used for grasping the knife. Therefore, the robot must be able to identify all graspable areas and choose the right one to plan the grasp and complete the task appropriately.

\begin{figure}[!t]
  \centering
  \includegraphics[width=\columnwidth]{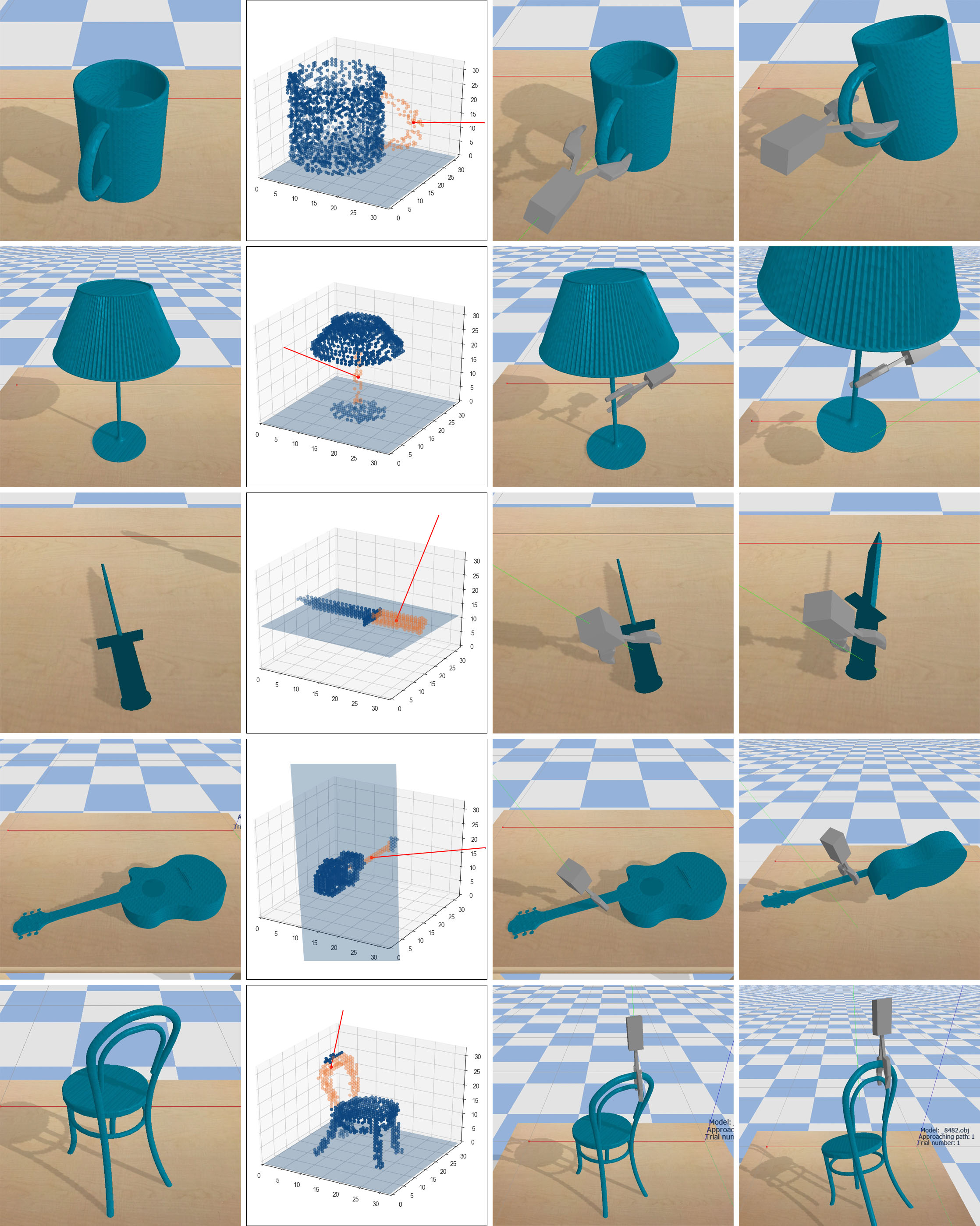}
  \caption{Examples of predicted grasp by the proposed Res-U-Net network on five sample objects. See the supplements for videos of the grasping trials.}
  \label{fig:object_detection}
  \vspace{-5mm}
\end{figure}

In this paper, we formulate the problem of grasp synthesis, i.e., finding a grasp configuration meeting a set of criteria for the specific grasping task \cite{bohg2013data}, as a learning problem. In particular, we propose a novel deep 3D Convolutional Neural Network (CNN) architecture to predict the graspable areas of the given object. We named it as Res-U-Net since it is built based on U-Net network architecture and residual blocks.  Our approach is designed to be robust and efficient to compute and use. Besides, we propose a method to find the best collision-free path to approach and grasp the object candidate using a parallel-plate robotic gripper.  The advantages of our approach over other state-of-the-art are:
\begin{itemize}
    \item Most of the recent works forced the robot to approach objects from above vertically (e.g., 3/4-DOF grasp \cite{mahler2017dex,morrison2018closing}). Such approaches simplify the problem of object grasping and cannot grasp planar objects, e.g., plates. In this paper, we propose a learning-based 6-DOF grasping approach that allows robots to approach the object from arbitrary directions.
    \item We show that our approach outperforms state-of-the-art architectures and enables a robot to pick up different objects with a success rate of 83.2\%. Fig.~\ref{fig:object_detection} shows five examples of our approach. Furthermore, we have tested the proposed method with a set of never-seen-before objects. Results showed that our approach generalizes well to new objects while generating only a small number of false predictions.
\end{itemize}

We extensively evaluate the performance of the proposed approaches in a simulation environment. The remainder of this paper is structured as follows. After reviewing related work, we discuss the proposed Res-U-Net architecture in Section~\ref{sec-affordance}. We then explain our ranking policy to select the best collision-free path for grasping the object in Section~\ref{sec-grasp}. Experimental results and discussion are given in Section~\ref{sec-experiments}, followed by conclusions in Section~\ref{sec-conclusions}.

\section{Related Work}
\label{sec-related}

Object grasping is one of the fundamental robotic tasks. Although an extensive survey is beyond the scope of this paper, we will review a few recent efforts.

Song \textit{et al.} \cite{song2015learning} developed a framework for estimating graspable parts of the objects from 2D images. Vahrenkamp \textit{et al.} \cite{vahrenkamp2016part} shown a system that could decompose novel object models by shape, local volumetric information, label them with semantic information, and plan the corresponding grasps. Kasaei \textit{et al.} \cite{kasaei2018towards} proposed an interactive open-ended learning approach to recognize and grasp novel objects in a human-centric environment. In another work, Kasaei \textit{et al.} developed a data-driven grasp approach to grasp household objects~\cite{kasaei2016object}. 

Over the past few years, extraordinary progress has been made in robotic applications with the emergence of deep learning approaches. Nguyen \textit{et al.} \cite{nguyen2016detecting} studied detecting 2D grasp affordance from RGB-D images by training a deep convolutional neural network. Kokic \textit{et al.} \cite{kokic2017affordance} utilized convolutional neural networks for encoding and detecting graspable parts of the object, class, and orientation to formulate grasp constraints. Mahler \textit{et al.} \cite{mahler2017dex} used a synthetic dataset to train a Grasp Quality Convolutional Neural Network (GQ-CNN) model, which can predict the probability of success of grasps from depth images. Choi \textit{et al.} \cite{choi2018learning} proposed a 3D convolutional neural network model to estimate grasp configuration using point cloud objects. Unlike Choi \textit{et al.}, our approach first predicts the graspable parts of the object, and then ranks them and plans to grasp the best part. 

Most of the grasp synthesis approaches mount an RGB-D camera on top of the workspace and use RGB or depth images to predict the grasp configuration as an oriented rectangle in the image frame (3-DOF). Therefore, such approaches necessitate the gripper pose to be perpendicular to the image plane, which leads to a set of drawbacks. The most important one is that picking up a planar object might be impossible given the top-down vertically approaching the object, and other constraints imposed by the robotic arm or task. In contrast to these approaches, our approach tackles the problem of predicting the 6-DOF grasp pose.

\section{Graspable Part Detection}
\label{sec-affordance}

Grasp synthesis denotes the formulation of a stable robotic grasp for a given object \cite{morrison2018closing}. In this paper, we formulate grasp synthesis as a cascaded approach: first predicting the graspable areas using Res-U-Net architecture, and then, choosing the best collision-free path to approach and grasp the object. The input to our grasp synthesis framework is a point cloud of an object extracted from a 3D scene using object segmentation algorithms such as \cite{kasaei2018perceiving,Sock_2017_ICCV}. A point cloud of an object, $O$, is represented as a set of points, $p_i : i \in \{ 1, \dots n\}$, where each point is described by their 3D coordinates $[x, y, z]$ and RGB information. In this work, we only use geometric information of the object and discard the color data. Therefore, we represent an object as a fixed occupancy grid of size $32 \times 32 \times 32$ voxels. The obtained representation is then used as the input to the Res-U-Net architecture to predict the graspable parts of the object, $g$, where $g \in O$. In the following subsections, we first discuss the architecture of two baseline networks and then describe Res-U-Net architecture in detail. We finally explain how to find the best grasp configuration and the collision-free path to grasp the target object.

\begin{figure}[!t]
  \centering
  \includegraphics[width=\columnwidth]{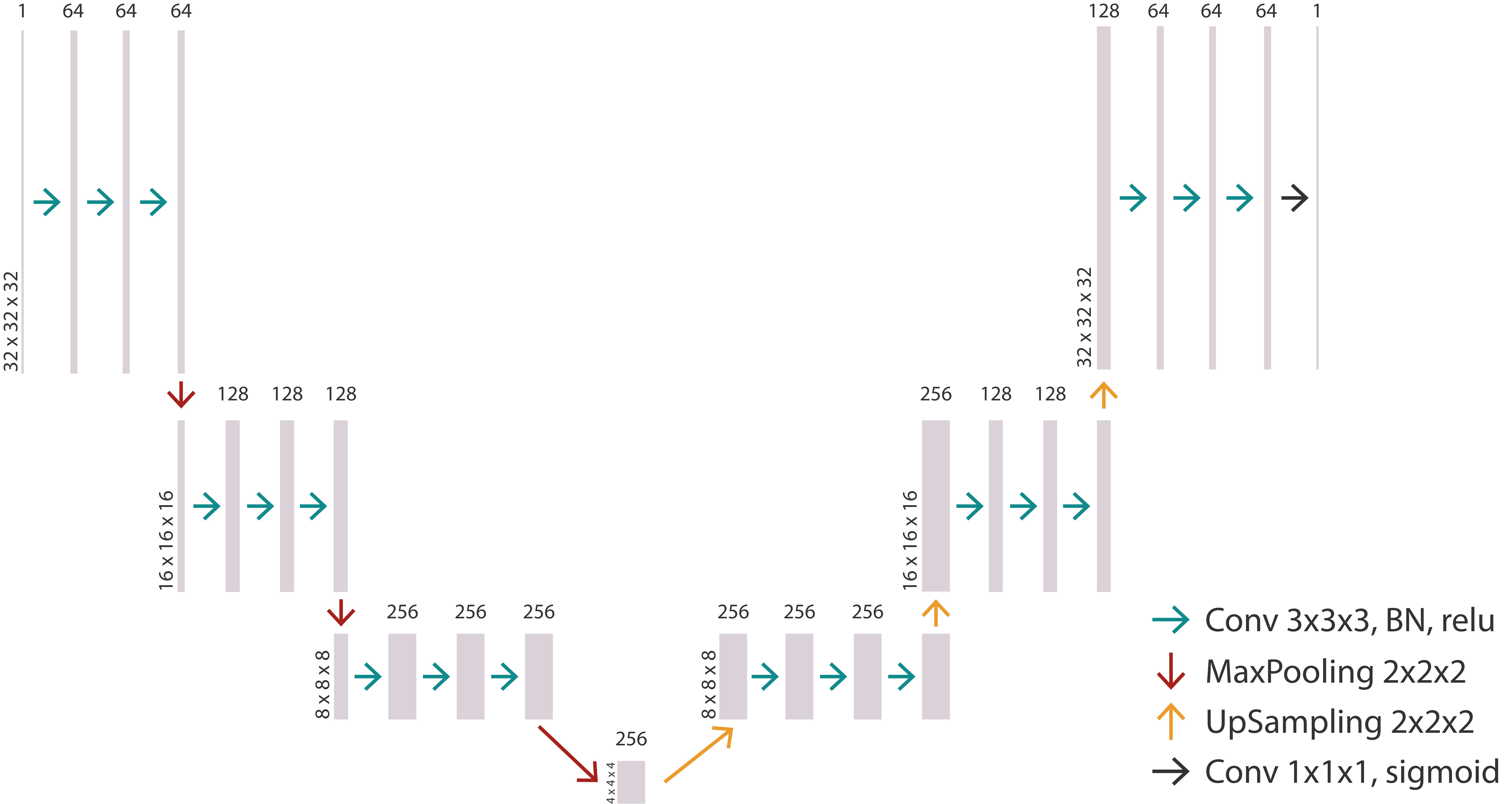}
  \caption{Structure of the encoder-decoder network. Each gray box stands for a multi-channel feature map. The number of channels is shown on the top of the feature map box. The shape of each feature map is denoted at the lower left of the box. The different color arrows represent various operations shown in the legend.}
  \label{f-norm}
  \vspace{-3mm}
\end{figure}

\begin{figure}[!b]
  \centering
  \includegraphics[width=\columnwidth]{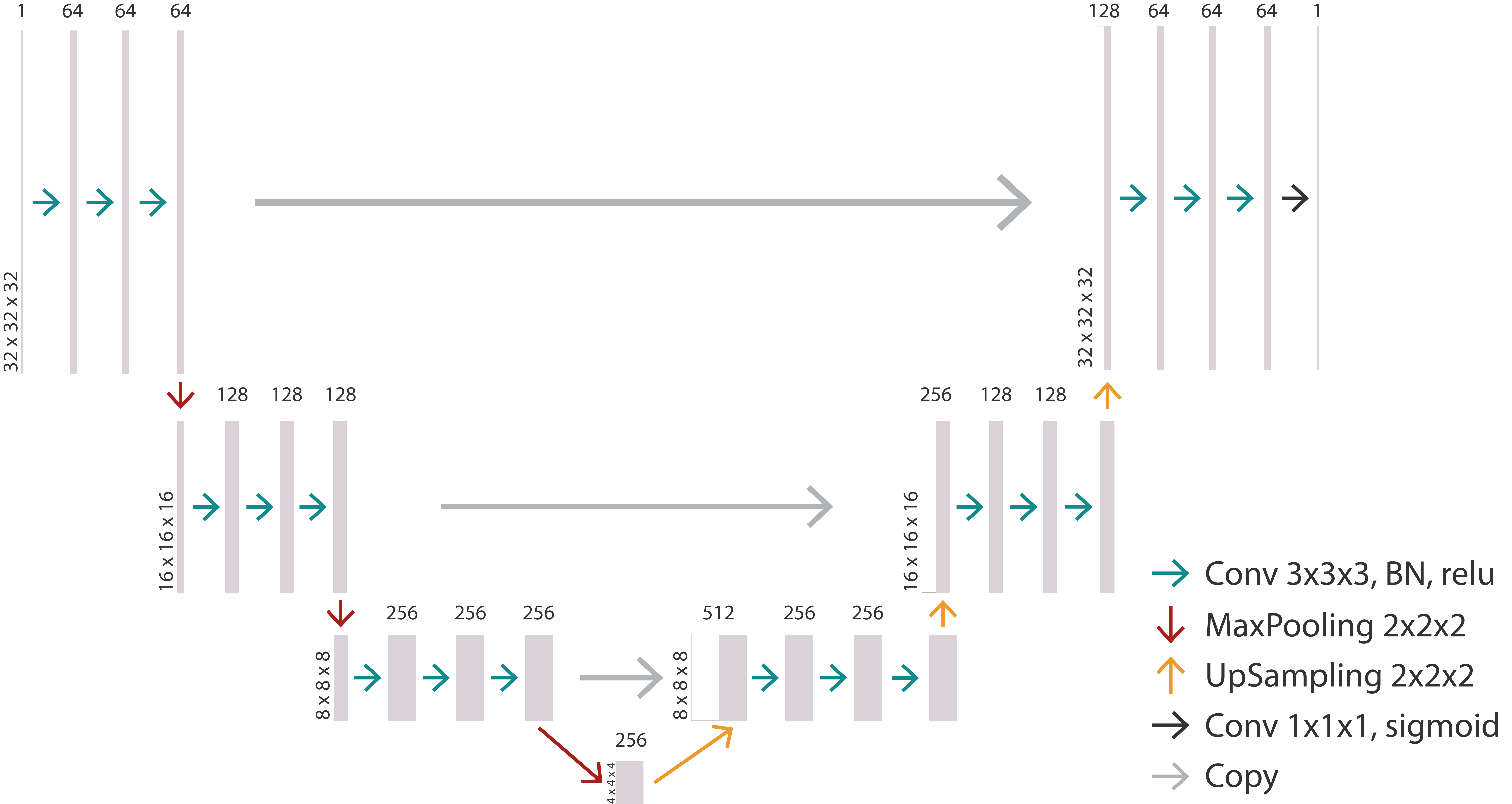}
  \caption{Structure of the U-Net. Compared to the encoder-decoder network, the last feature map of each layer in the encoder part is copied and concatenated to the first feature map of the same layer in the decoder part.}
  \label{f-unet}
  \vspace{-3mm}
\end{figure}

\begin{figure*}[!t]
  \centering
  \includegraphics[width=0.825\linewidth]{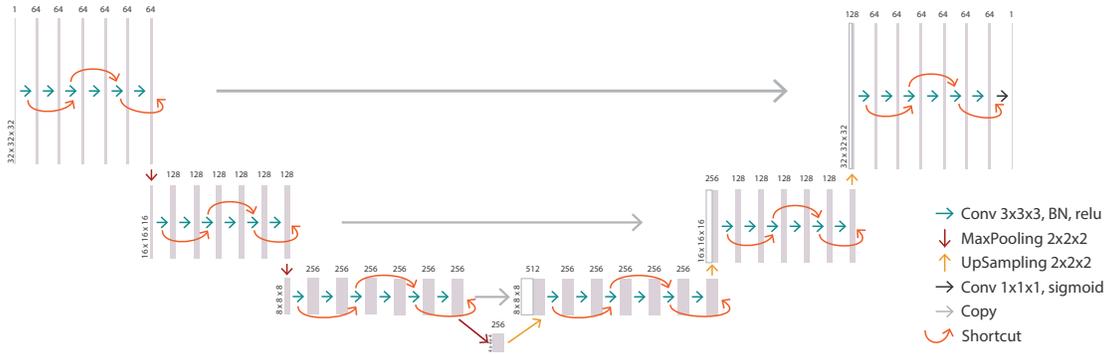}
  \caption{Structure of the proposed Res-U-Net. Compared to the U-Net, we replace the residual blocks with 3D convolutional layers and skipping over layers. This skipping over layers effectively simplifies the network and speeds learning by reducing the impact of vanishing gradients.}
  \label{f-resunet}
  \vspace{-5mm}
\end{figure*}

\subsection{Baseline Networks}

To make our contribution transparent, we build two baselines, including autoencoder architecture~\cite{yang2016object}, and U-Net network~\cite{ronneberger2015u} and highlight the similarities and differences between these approaches and our Res-U-Net. All the networks contain two essential parts: one is the encoder network, and the other is the decoder network.

The architecture of the encoder-decoder network~\cite{yang2016object} is depicted in Fig.~\ref{f-norm}. This architecture is the lightest one among the selected architectures in terms of the number of parameters and computation time, making the network easier and faster to learn. The encoder part of this network has nine 3D convolutional layers (all of them are $3 \times 3 \times 3$), and each of them is followed by batch normalization and ReLU layer. At the end of each encoder layer, there is a 3D max-pooling layer of $2 \times 2 \times 2$ to produce a dense feature map. Each encode layer is corresponding to a decoder layer. It also has nine 3D convolutional layers. The difference is that instead of having 3D max-pooling layers, at the beginning of each layer, an up-sampling layer is utilized to produce a higher resolution of the feature map. Besides, a $1 \times 1 \times 1$ convolutional layer and a sigmoid layer is attached after the final decoder to reduce the multi-channels to $1$.

The architecture of U-Net \cite{ronneberger2015u} is shown in Fig.~\ref{f-unet}. The basic structure of the U-Net and the described encoder-decoder network are almost the same. The main difference is that, in U-Net architecture, the dense feature map is first copied from the end of each encoder layer to the beginning of each decoder layer. Then the copied layer and the up-sampled layer are concatenated.

\subsection{Proposed Res-U-Net Network}

The architecture of our approach is illustrated in Fig.~\ref{f-resunet}. As shown in this figure, the network architecture is a combination of U-Net and residual network \cite{he2016deep}. We, therefore, call this network Res-U-Net. We come up with this architecture to retain more information from the input layer and dig more features, inspired by the residual network \cite{he2016deep}. Compared to the U-Net, we replace the residual blocks with 3D convolutional layers and skipping over layers. The primary motivation is to overcome the vanishing gradients problem by reusing activation from a previous layer until the adjacent layer learns its weights. The network can go deeper using the residual blocks, since it simplifies the network by considering fewer layers in the initial training stages. The encoder and decoder parts are jointly trained to minimize the average reconstruction loss $\mathcal{L}(g^{'}, g)$ between the predicted graspable areas, $g^{'}$, and the ground truth areas, $g$, over a training set. 

\begin{figure*}[thpb]
  \centering
  \includegraphics[width=0.965\linewidth]{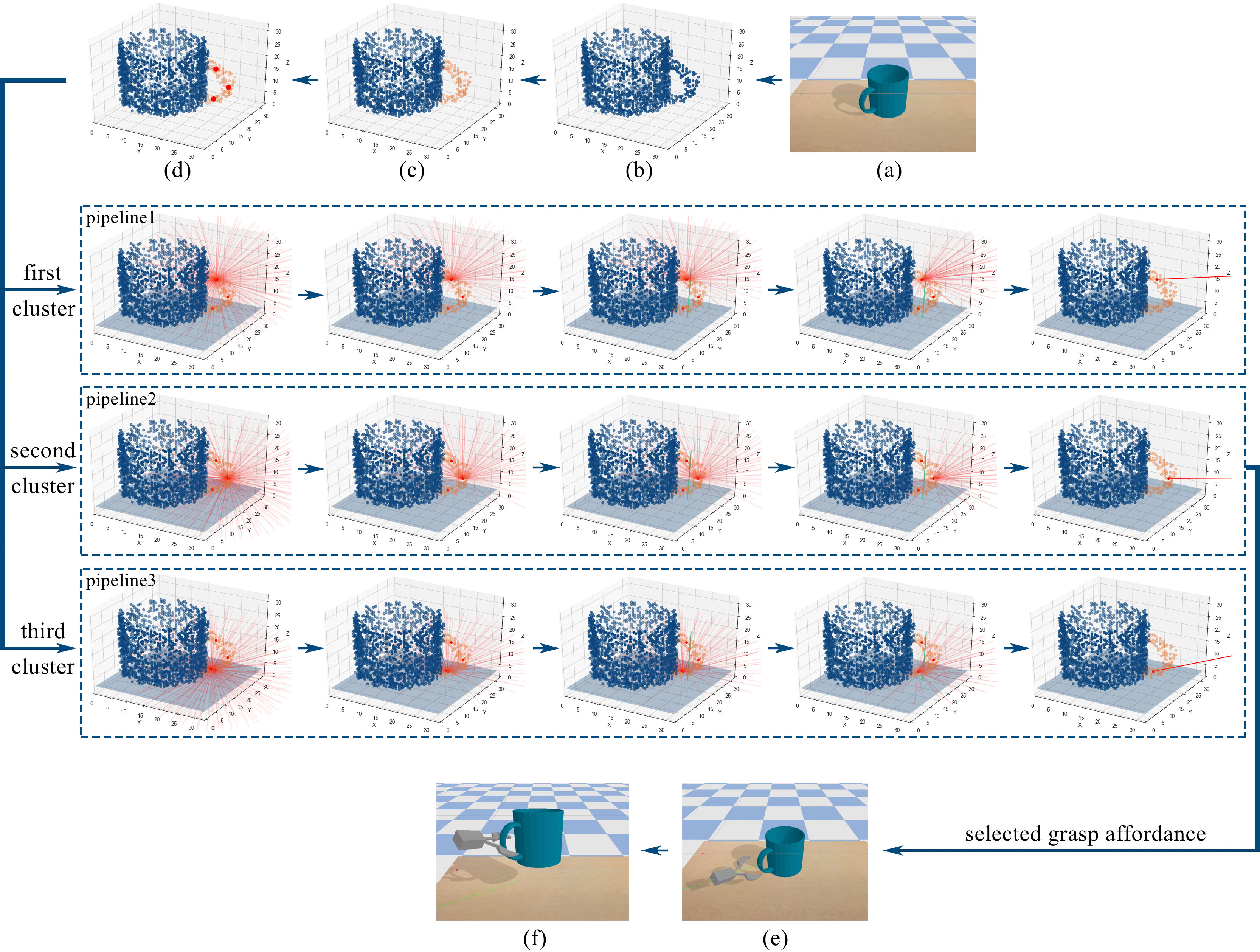}
  \caption{An illustrative example of the proposed object grasping approach: (\emph{a}) a \emph{Mug} object in our simulation environment; (\emph{b}) point cloud of the object; (\emph{c}) feeding the point cloud to Res-U-Net for detecting the graspable part of the object (highlighted by orange color); (\emph{d}) the predicted area is then segmented into three clusters using the K-means algorithm. The centroid of each cluster is considered as a graspable point. Then, the point cloud of the object is further processed in three pipelines to find out an appropriate 6-DoF grasp configuration for each graspable point. In particular, inside each pipeline, a set of approaching paths is first generated based on the Fibonacci sphere (shown by red lines) and the table plane information (shown by a dark blue plane); we then eliminate those paths that go through the table plane. Afterward, we find the principal axis of the graspable part by performing PCA analysis (the green line shows the main axis), which is used to define the goodness of each approaching path. The best approaching path is finally detected and (\emph{e}) used to perform grasping; (\emph{f}) this snapshot shows a successful example of grasp execution.}
  \label{f-pipeline}
  \vspace{-5mm}
\end{figure*}

\section{Ranking Collision-free Paths for Grasping}
\label{sec-grasp}

To discuss the problem better, we provide a representative example of the proposed approach in Fig.~\ref{f-pipeline}. As shown in Fig.~\ref{f-pipeline} (\textit{a}), we assume that a given object is laying on a planar surface. The object is then extracted from the scene and fed to the Res-U-Net (see \emph{b-c}). After detecting the graspable area of the given object, the point cloud of the object is further processed to determine an appropriate 6-DOF grasp configuration (i.e., the position and the orientation of end-effector in 3D space). In particular, the predicted graspable part of the object is first segmented into $m$ clusters using the K-means algorithm, where $m$ is defined based on the size of the graspable part of the object and robot's griper. The centroid of each cluster indicates one grasp candidate (see Fig.~\ref{f-pipeline} (\emph{d})). Each centroid is considered as the desired pose of the approaching path. We create a pipeline for each grasp candidate and process the object further to define the starting pose of the collision-free approaching path. Inside each pipeline, we generate a Fibonacci sphere by putting the center of the sphere at the grasp candidate and then randomly select $N$ points on the sphere. We then define $N$ linear approaching paths by calculating lines using selected points and the grasp candidate point (i.e., the center of the sphere). In our current setup, $N$ has been set to 256 points shown by red lines Fig.~\ref{f-pipeline}. In this study, we use the following procedures to define the best collision-free approaching path:

\begin{itemize}
    \item \textbf{Discard infeasible paths:}  by considering the table information, we remove infeasible approaching paths. Particularly, those paths that their starting point is under the table or the paths that gripper collides with the table before reaching the object (\emph{see the second image in each pipeline}).
    
    \item \textbf{Compute the main axis of the predicted graspable part using Principal Component Analysis (PCA):} The axis with maximum variance is considered as the main-axis (\emph{shown by a green line in the third image of each pipeline}).
    
    \item \textbf{Rank each of the approaching path:} 
    we propose the following equation to rank each of the remaining paths:
    \begin{equation}
         score = 2 \frac{\pi - a}{\pi} \sum_{i=1}^{n} min(1, \frac{1}{d^2+\epsilon})
    \end{equation}
where $n$ represents the number of points of the object, $d$ stands for the distance between the specific approaching path and one of the points in a point cloud model, \(\epsilon\) is equal to $0.01$, and $a$ is the angle between approaching path line and the main axis of the graspable part of the object, ranging from $0$ to \(\frac{\pi}{2}\). Since \cite{balasubramanian2010human} has shown that humans tend to grasp object orthogonal to the principal axis, we then calculate \((2 * \frac{\pi - a}{\pi})\) in the formula to reduce the score when the path is orthogonal to the principal axis. The lower score means the distances between the approaching path to all points of the objects are farther. Therefore, the path with the lowest score is selected as a final approaching path for each grasp point candidate. The approaching paths with scores' influence are shown as the fourth image in each pipeline. It is visible that all paths with deeper color represent proper approaching paths. Finally, the best approaching path is selected as the approaching path for the given grasp point (\emph{last figure in each pipeline}).

\end{itemize}

After calculating the best collision-free approaching path, we instruct the robot to follow the path. Towards this end, we first transform the approaching path from object frame to world frame and then dispatch the planned trajectory to the robot to be executed (Fig.~\ref{f-pipeline} (\emph{e} and \emph{f})). It is worth mentioning that in some situations, fingers of the gripper touch the table (which stops the gripper from moving forward). To handle this point, we do slight roll rotation on the gripper to find a better angle between gripper and table to keep gripper moving forward. An illustrative example of the proposed object grasping approach is depicted in Fig.~\ref{f-pipeline}.

\section{Experimental Results}
\label{sec-experiments}

A set of experiments was carried out to evaluate the proposed approach. In this section, we describe our experimental setup and discuss the obtained results.

\subsection{Dataset and Evaluation Metrics}

In these experiments, we mainly used a subset of ShapeNetCore \cite{shapenet2015} containing 500 models from five categories, including \emph{Mug, Chair, Knife, Guitar}, and \emph{Lamp}. 
For each category, we randomly selected object models and converted them into complete point clouds with the pyntcloud package. We then shifted and resized the point cloud data and turn them into a $32 \times 32 \times 32$ array as the input size of networks.  

\begin{figure}[!b]
  \vspace{-4mm}
  \centering
  \includegraphics[width=0.8\linewidth]{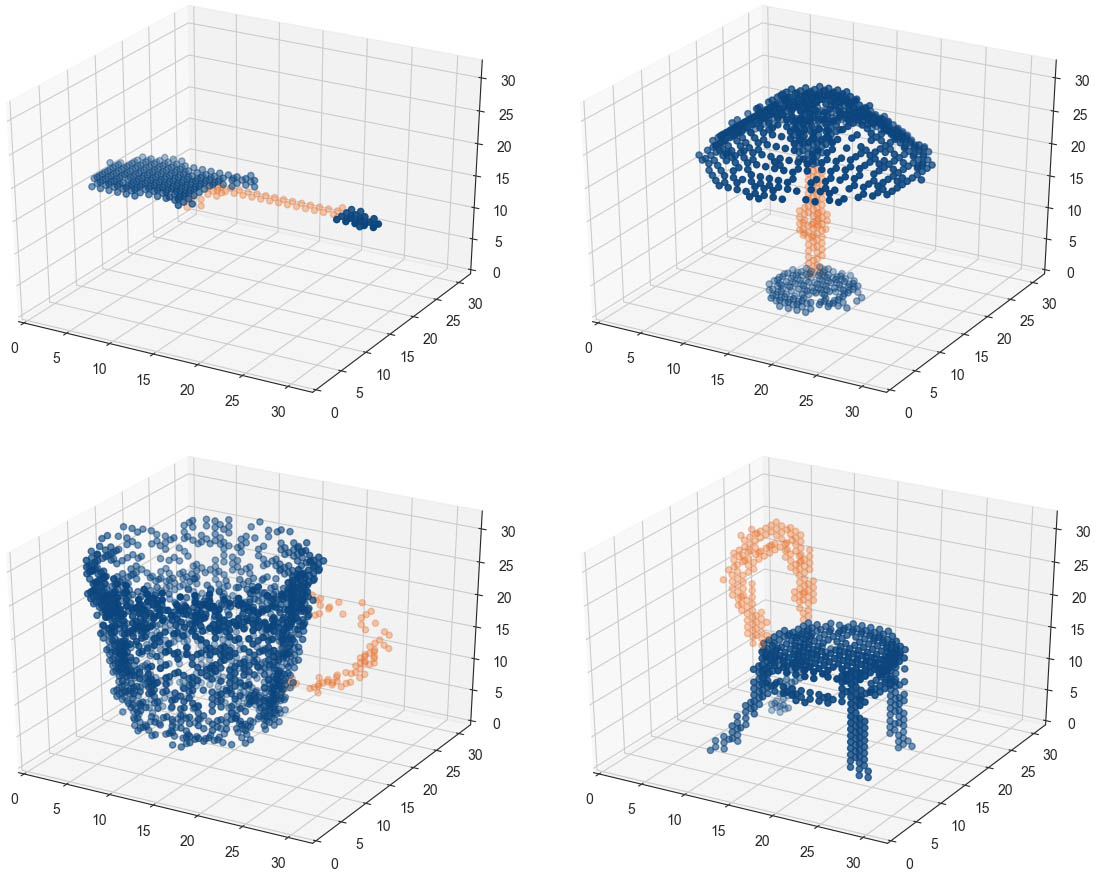}
  \caption{Examples of labeling graspable parts for four objects: point cloud of the object is shown by dark blue and graspable parts are highlighted by orange color.}
  \label{f-mug}
\end{figure}

We manually labeled graspable parts for each object to provide ground truth data. In particular, part annotations are represented as point labels.  A set of examples of labeled graspable parts for different objects is depicted in Fig.~\ref{f-mug} (graspable parts are highlighted by orange color). It should be noted that we augmented the dataset by rotating the point clouds along the z-axis for $90$, $180$, and $270$ degrees and flipping the point clouds vertically and horizontally from the top view to augment the training and validation data.  We obtained $2580$ training, $210$ validation and $210$ test data for evaluation. For researchers who want to delve into this area, we make our dataset publicly available at: 
\href{http://github.com/yikun-li/pc-3d-grasp-ds.git}{\cblue{\footnotesize \textbf{\texttt{http://github.com/yikun-li/pc-3d-grasp-ds.git}}}}.

We mainly used Average Intersection over Union (IoU) as the evaluation metric. We first computed IoU for each part of the object. Afterward, for each category, IoU was computed by averaging per part IoU across all parts of all objects. To evaluate the grasping part, we used \textit{success rate} metric, which is defined as the ratio of successful grasps to all performed grasp experiments.

\subsection{Training}

All the proposed networks were trained from scratch through \texttt{RMSprop} optimizer with the $\rho$ setting to $0.9$. We initially set the learning rate to $0.001$. If the validation loss does not decrease in 5 epochs, the learning rate is decayed by multiplying the square root of $0.1$ until it reaches the minimum learning rate of \num{0.5e-6}. The \texttt{binary cross-entropy} loss was employed in training, and the batch size was set to 16. We mainly used Python and Keras library in this study. The training process took around two days on our NVIDIA Tesla K40m GPU, depending on the complexity of the network.

\begin{figure}[!t]
  \centering
  \includegraphics[width=1\linewidth]{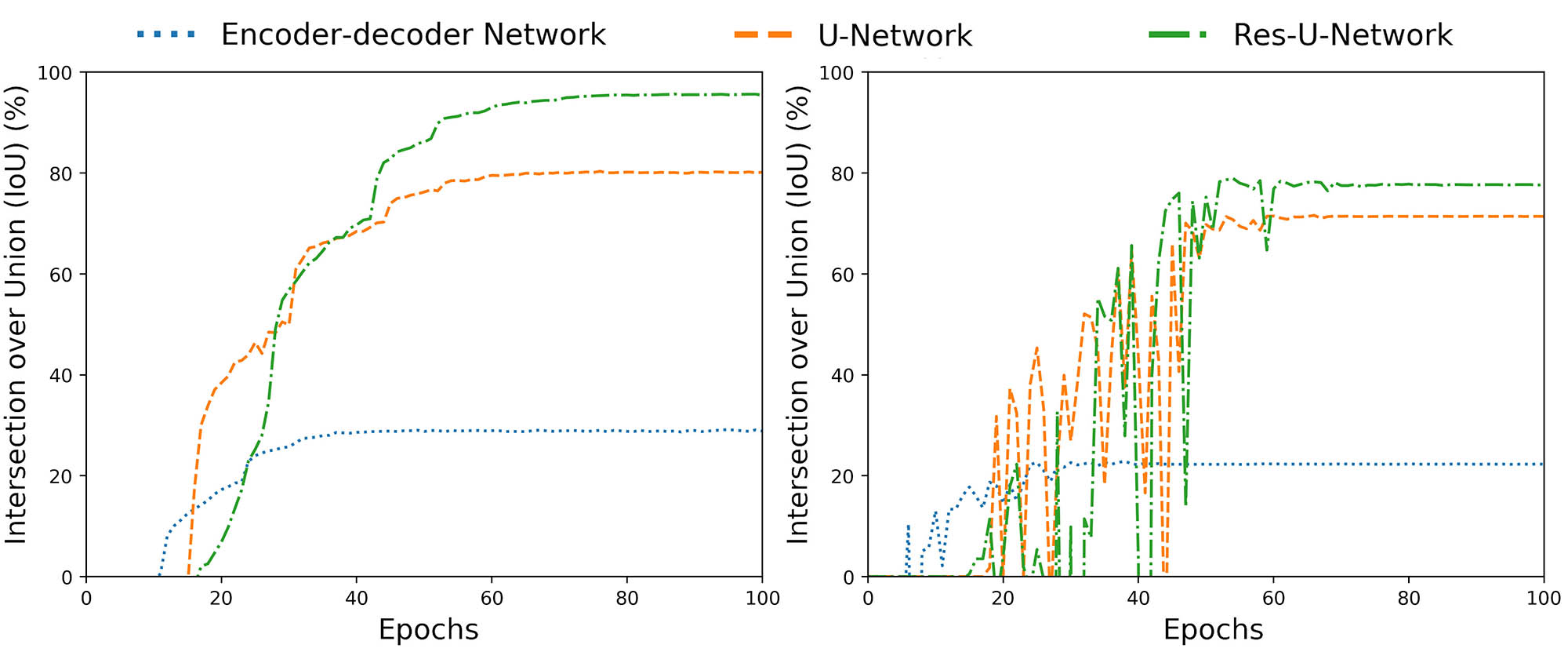}
  \caption{Learning curves of different approaches during (\textit{left}) training phase, and (\textit{right}) validation phase as a function of IoU vs. epochs.}
  \label{f-iou_over_epochs}
  \vspace{-5mm}
\end{figure}

\subsection{Graspable Part Prediction}

\begin{figure}[!b]
  \vspace{-5mm}
  \centering
  \includegraphics[width=0.78\linewidth]{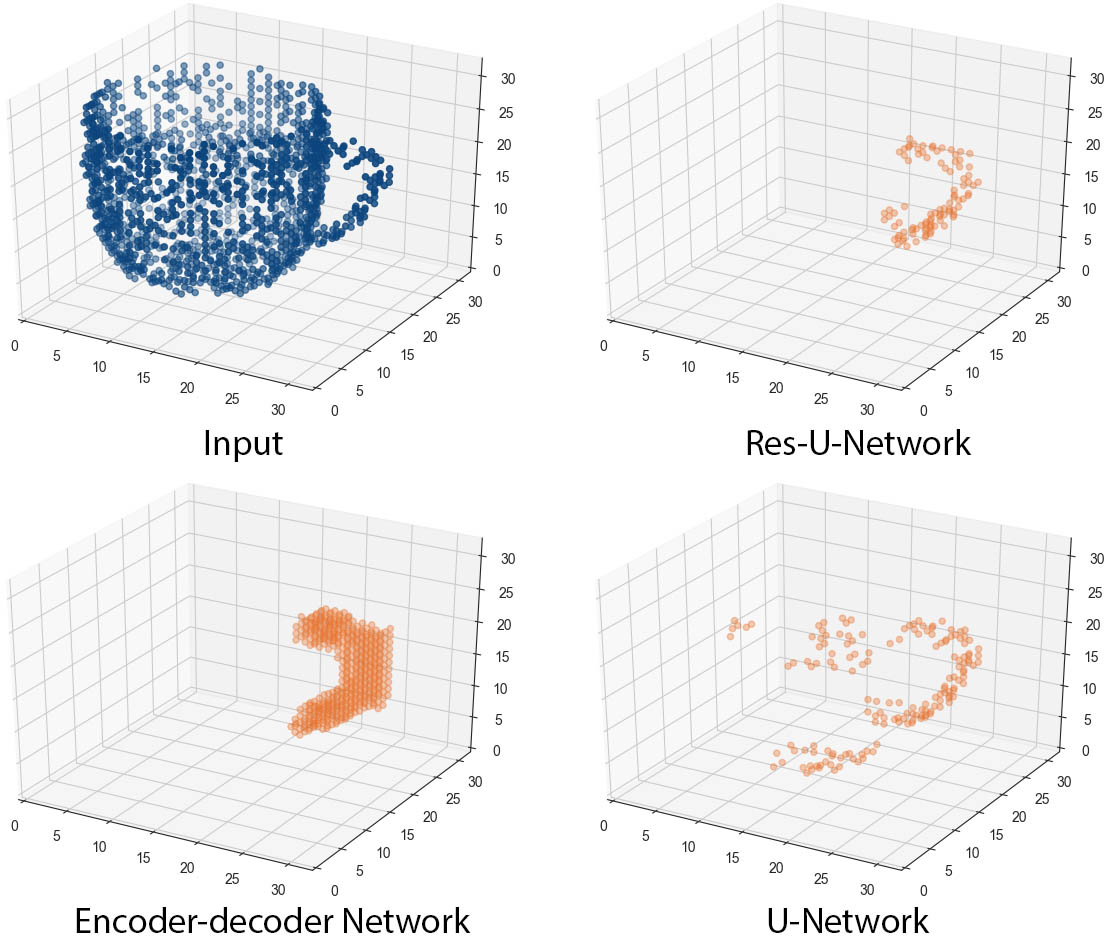}
  \caption{An example of predicting graspable parts of a mug object by the proposed Res-U-Net, the encode-decoder network, and the U-Net network.}
  \label{f-result_comparison}
\end{figure}

Fig.~\ref{f-iou_over_epochs} shows the progress of the proposed networks over 100 epochs. By comparing all the experiments, it is visible that the encoder-decoder network performs much worse than the other approaches. In particular, the final IoU of the encoder-decoder network is 28.9\% and 22.3\% on training and validation data, respectively. The U-Net performs much better than the encoder-decoder network, in which its final IoU is 80.1\% on training and 71.4\% on validation data. The proposed Res-U-Net architecture outperforms the others by a large margin. The final IoU of Res-U-Net is 95.5\% and 77.6\% on training and validation data, respectively. Notably, in the case of training, it is 15.4 percentage points (p.p.) better than U-Net and  66.6 p.p. better than the encoder-decoder network, in the case of validation, it is 6.2 p.p., and 55.3 p.p. better than U-Net and encoder-decoder network respectively.
Fig.~\ref{f-result_comparison} shows an example of detecting graspable parts of a mug object by different networks.

\subsection{Grasping Results}

We evaluated our grasp methodology using a simulated robot. In particular, we built a simulation environment to verify the capability of the proposed grasp approach. The simulation was developed based on the Bullet physics engine. We only considered the end-effector pose $(x , y , z , roll , pitch , yaw)$ to simplify the complexity and concentrate on evaluating the proposed approach. 

We designed a grasping scenario that the simulated robot first grasps the object and then picks it up to a certain height to see if the object slips due to the bad grasp configuration. 
If the robot can complete the task, the grasp is considered as a successful grasp.
In this experiment, we randomly selected 50 different objects for each of the five mentioned categories. In each experiment, we randomly placed the object on the table region and also rotate it along the z-axis. It is worth to mention that all test objects were not used for training the neural networks. We achieved a grasp success rate of 83.2\% (i.e., 208 success out of 250 trials). The detailed outcomes of the experiments are summarized in Table~\ref{table-grasp}.
Fig.~\ref{fig:object_detection} shows the grasp detection results of five example objects.  A video of this experiment is available online at \href{http://youtu.be/5_yAJCc8owo}{\cblue{\footnotesize \textbf{\texttt{http://youtu.be/5\_yAJCc8owo}}}}.

\begin{table}[!b]
\vspace{-3mm}
\caption{Grasp success rate on five categories}
\label{table-grasp}
\begin{center}
\resizebox{0.78\columnwidth}{!}{
\begin{tabular}{ccc}
\hline
Category & Success rate & Success / Total \\
\hline
Mug & 0.86 & 43 / 50 \\
Chair & 0.80 & 40 / 50 \\
Knife & 0.84 & 42 / 50 \\
Guitar & 0.80 & 40 / 50 \\
Lamp & 0.86 & 43 / 50 \\
\hline
Average & 0.832 & 208 / 250 \\
\hline
\end{tabular}
}
\end{center}
\end{table}

\begin{figure}[!t]
 \centering
 \includegraphics[width=1\columnwidth]{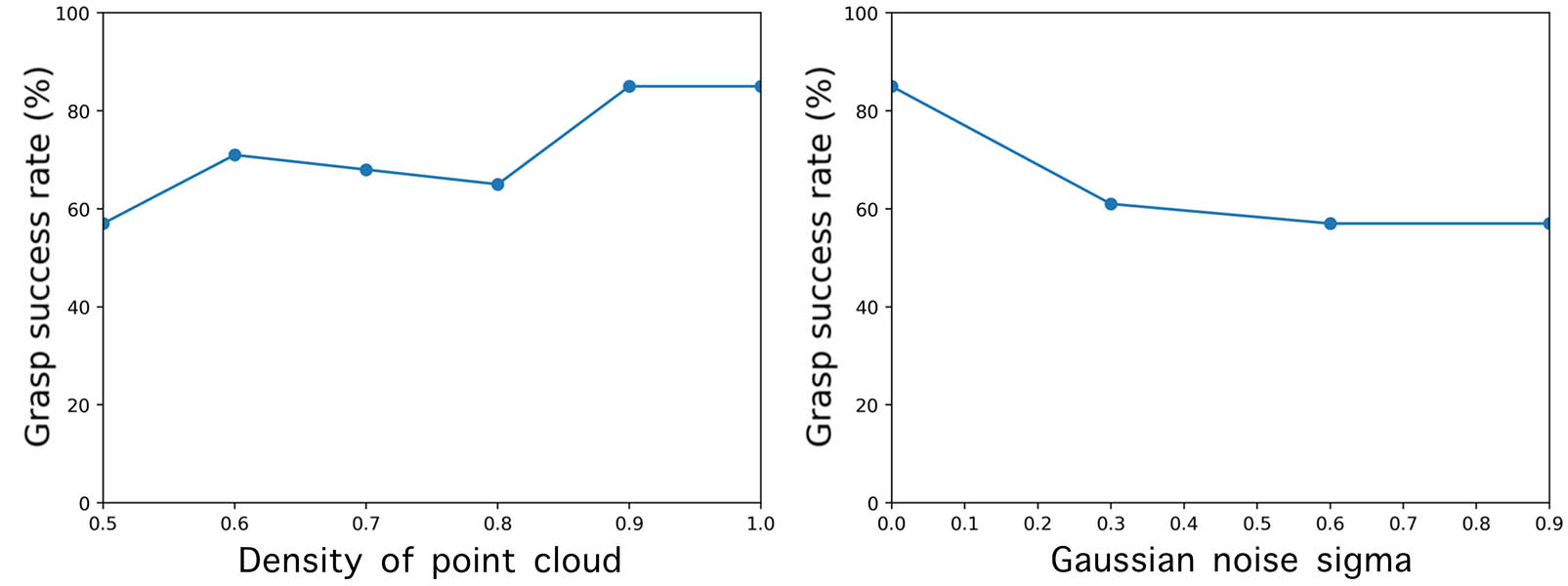} 
 \caption{The robustness of the Res-U-Net to (\textit{left}) varying point cloud density, and (\textit{right}) different level of Gaussian noise.}
 \label{fig:robustness_simulation}
 \vspace{-5mm}
\end{figure}

Two sets of experiments were conducted to examine the robustness of the proposed approach concerning varying point cloud density and Gaussian noise. Particularly, in the first set of experiments, the original density of training objects was kept, and the density of testing objects was reduced (downsampling) from 1 to 0.5. In the second set of experiments, nine levels of Gaussian noise were added to the test data. The results are summarized in Fig.~\ref{fig:robustness_simulation}.

From experiments of reducing the density of test data (i.e., Fig.~\ref{fig:robustness_simulation} (\textit{left}), it was found that our approach is robust to low-level downsampling, i.e., with 0.9 point density, the success rate remains the same. In the mid-level downsampling resolution (i.e., point density between 0.6 and 0.8), the grasp success rate dropped around 20\%. It can be concluded from Fig.~\ref{fig:robustness_simulation} (\textit{left}) that when the level of downsampling increases to 0.5, the grasp success rate dropped to 57\% rapidly.

In the second round of experiments, Gaussian noise was independently added to the $X$, $Y$, and $Z$ axes of the given test object. As shown in Fig.~\ref{fig:robustness_simulation} (\textit{right}), performance decrease when the standard deviation of the Gaussian noise increases. In particular, when we set the sigma to $0.3$, $0.6$, and $0.9$, the success rates are dropped to $61\%$, $57\%$, and $57\%$, respectively.

\begin{figure}[!t]
  \centering
  \includegraphics[width=\columnwidth]{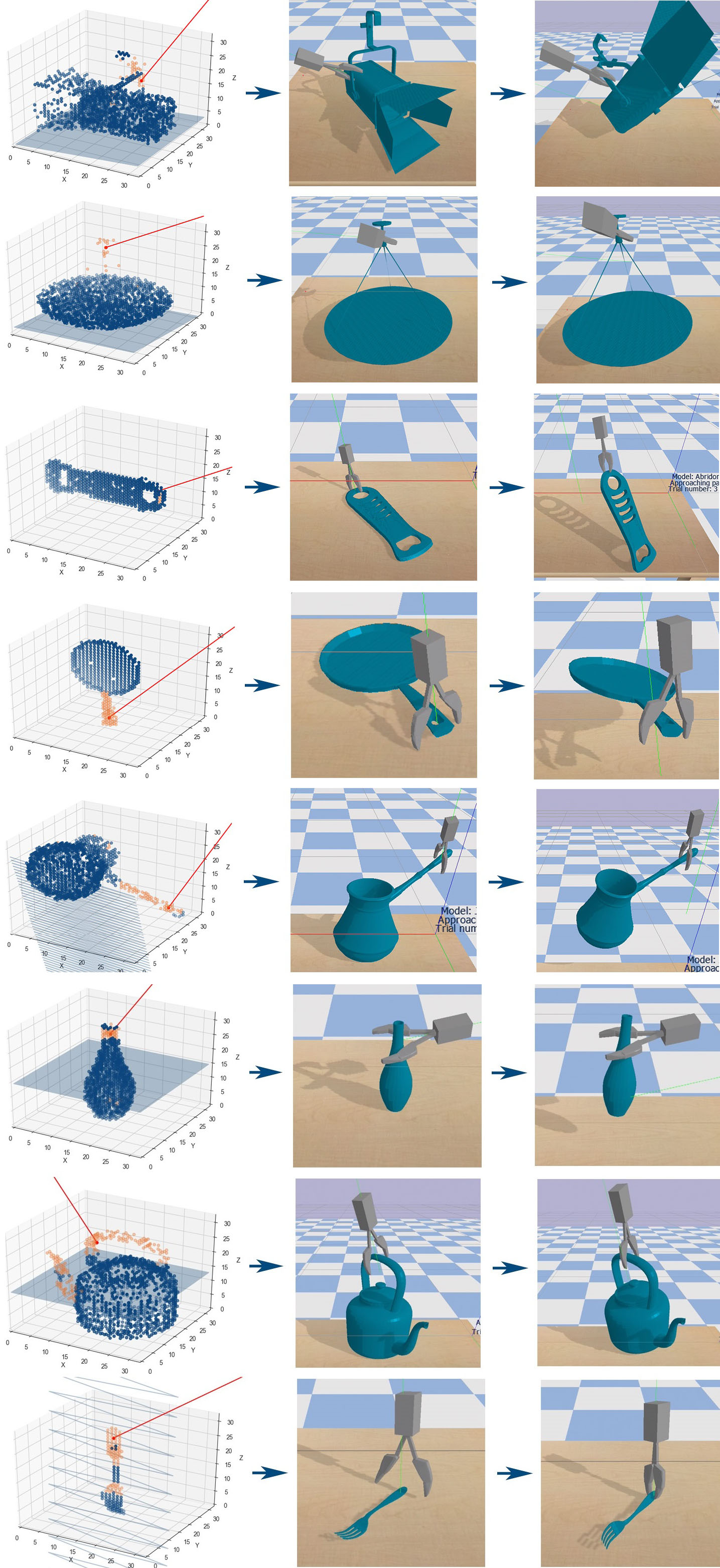}
  \caption{Examples of grasping unknown objects: our approach is able to predict an appropriate graspable part of each object, and find a collision-free path to approach and pick up the object successfully.}
  \label{f-extra}
  \vspace{-5mm}
\end{figure}

Our approach was trained to grasp five object categories. In this experiment, we examined the performance of our approach by a set of 50 never-seen-before objects. In most cases, the robot could detect an appropriate grasp configuration for the given object and complete the grasping scenario. This observation showed that the proposed Res-U-Net could use the learned knowledge to grasp most of the unknown objects correctly. In particular, a never-seen-before object that is similar to one of the known ones (i.e., they are familiar) can be grasped similarly. Fig.~\ref{f-extra} shows the steps taken by the robot to grasp a set of unknown objects in our experiments.

In both experiments (i.e., grasping known and unknown objects), we have encountered three types of failure modes. First, Res-U-Net may fail to predict an appropriate part of the object for grasping. Second, grasping may fail because of the collision between the gripper, object, and table. It could also fail because the predicted graspable part was too small to grasp the target object, or the graspable area was too large to fit in the robot's gripper (e.g., the body of Mug). In some cases, it happened if the object is too big or slippery (e.g., Chair and Lamp). The last case of failure was when the finger of the gripper is tangent to one of the object's surfaces. In such cases, although the graspable part of the object was correctly predicted, the robot pushed the object instead of grasping it.

Another set of experiments was performed to estimate the execution time of the proposed approach. Three components mainly make the execution time: perception, graspable part prediction, and finding the best collision-free approaching path. We measured the run-time for ten instances of each. Perception of the environment and converting the point cloud of the object to appropriate voxel-based representation (on average) took $0.15$ seconds. Graspable part prediction by Res-U-Net required $0.13$ seconds on average, and finding suitable grasp configuration demanded another $1.32$ seconds. Therefore, finding a complete grasp configuration for a given object on average took about $1.60$ seconds.

\section{Conclusion and Future Work}
\label{sec-conclusions}

In this paper, we present a novel deep convolutional neural network named Res-U-Net to detect graspable parts of 3D Objects. The point cloud of the object is further processed to determine an appropriate grasp configuration for the predicted graspable parts of the object. To validate our approach, we built a simulation environment and conducted an extensive set of experiments. Results show that the overall performance of the proposed Res-U-Net is clearly better than the best results obtained with the U-Net and Autoencoder approaches. We also test our approaches by a set of never-seen-before objects. It was observed that, in most of the cases, our approach was able to detect graspable parts of the objects correctly and perform the proposed grasp scenario successfully. In the continuation of this work, we plan to evaluate the proposed approach in clutter scenarios, such as clearing a pile of toy objects. We would also like to investigate the possibility of considering Res-U-Net for task-informed grasping scenarios.

\bibliography{bibliography}
\bibliographystyle{IEEEtran}

\end{document}